\documentclass{article}

\usepackage[preprint]{neurips_data_2021}

\usepackage[utf8]{inputenc} %
\usepackage[T1]{fontenc}    %
\usepackage{hyperref}       %
\usepackage{url}            %
\usepackage{booktabs}       %
\usepackage{amsfonts}       %
\usepackage{nicefrac}       %
\usepackage{microtype}      %
\usepackage{xcolor}         %
\usepackage{graphicx}
\usepackage{subcaption}

\usepackage{wrapfig,lipsum,booktabs}

\title{CrossedWires: A Dataset of Syntactically Equivalent but Semantically Disparate Deep Learning Models}

\author{%
  Max Zvyagin, Thomas Brettin, Arvind Ramanathan \\
  Argonne National Laboratory\\
  Lemont IL 64309 \\
  \texttt{\{mzvyagin,brettin,ramanathana\}@anl.gov}\\
   \And
   Sumit K. Jha \\
   Computer Science Department \\
   University of Texas at San Antonio \\
   \texttt{sumit.jha@utsa.edu} 
   }

\begin{document}

\maketitle

\begin{abstract}
The training of neural networks using different deep learning frameworks may lead to drastically differing accuracy levels despite the use of the same neural network architecture and identical training hyperparameters such as learning rate and choice of optimization algorithms. Currently, our ability to build standardized deep learning models is limited by the availability of a suite of neural network and corresponding training hyperparameter benchmarks that expose differences between existing deep learning frameworks. In this paper, we present a living dataset of models and hyperparameters, called CrossedWires, that exposes semantic differences between two popular deep learning frameworks: PyTorch and Tensorflow. The CrossedWires dataset currently consists of models  trained on CIFAR10 images using three different computer vision architectures: VGG16, ResNet50 and DenseNet121 across a large hyperparameter space. Using hyperparameter optimization, each of the three models was trained on 400 sets of hyperparameters suggested by the HyperSpace search algorithm. The CrossedWires dataset includes PyTorch and Tensforflow models with test accuracies as different as 0.681 on syntactically equivalent models and identical hyperparameter choices. The 340 GB dataset and benchmarks presented here include the performance statistics, training curves, and model weights for all 1200 hyperparameter choices, resulting in 2400 total models. The CrossedWires dataset provides an opportunity to study semantic differences between syntactically equivalent models across popular deep learning frameworks. Further, the insights obtained from this study can enable the development of algorithms and tools that improve reliability and reproducibility of deep learning frameworks. The dataset is freely available at \url{https://github.com/maxzvyagin/crossedwires} through a Python API and direct download link. 
\end{abstract}

\section{Introduction}

\begin{figure}
    \centering
    \includegraphics[scale=0.3]{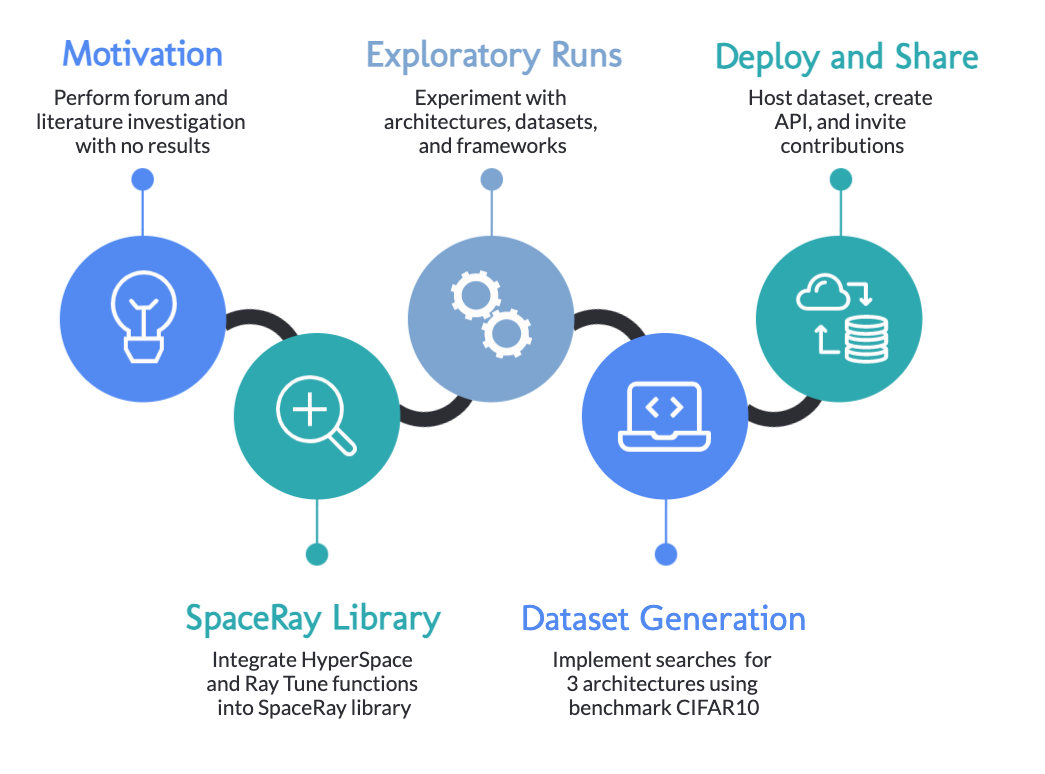}
    \caption{An overview of the CrossedWires dataset: We studied the problem of syntactically identical models producing different test accuracies using formal publications and informal anecdotal evidence on forums. We built the SpaceRay library for easily parallelizing hyperparameter search using multiple GPU nodes. We performed $\mathcal{O}(10^4)$ GPU hours of exploratory runs to determine the sensitive hyperparameters and their ranges. Finally, we created the CrossedWires dataset of 1200 pairs of models and built the Python API for efficiently accessing the dataset.}
    \label{fig:infographic}
\end{figure}

Deep learning and neural network  techniques \citep{10.1145/3448250} are ubiquitous within AI applications that are deployed at scale across various domains (speech \citep{xiong2017toward, saon2015ibm, wang2018supervised, amodei2016deep, huang2014deep}, language translation \citep{young2018recent, vaswani2018tensor2tensor, bahdanau2014neural}, self-driving cars \citep{maqueda2018event, ramos2017detecting, kulkarni2018traffic}, image search and recognition \citep{krizhevsky2012imagenet, he2015delving, abbas2021classification, sun2015deepid3}, information retrieval \citep{lam2015combining, liu2015representation, yan2016learning}, etc.). However, the choice of deep learning frameworks and consequently the hardware (or accelerator; e.g., GPU, TPU, data-flow architectures, etc.) platform used to implement, train, and deploy  neural network architectures has an immediate impact on their performance. In specific, syntactically equivalent neural network implementations (across deep learning frameworks) can have a large variance in model accuracies despite controlling for architectures, hyperparameter settings, and training/test data. Currently, only anecdotal evidence of such variance exists~\citep{tensorflow-github-issue, tensorflow-github-issue2, pytorch-discuss-post, stack-overflow}, which makes it challenging to study such variations in a systematic manner. This often results in poor reproducibilty of deep learning models, often varying even across different versions of the same  library \citep{renard2020variability}. 

Given the wide range of applications that deep learning models are being deployed on, including safety-critical systems \citep{8569637, 8897630, henne2020benchmarking} and scientific/experimental workflows (e.g., as surrogate models \citep{9355293, christensen2021orbnet, kashinath2021physics, jumper2021highly, Townshend1047}, or optimal experimental design strategies \citep{CAO2021100295, king1996structure, sparkes2010towards}), there is an immediate need to develop datasets that can quantify the intrinsic variability in model performance based on the choice of framework selected alone as well as how this may impact the choice of performance optimization on a specific hardware platform. The end users of deep learning frameworks naturally expect the accuracy of the models to depend only on the model architecture, training algorithm and the choice of hyperparameters. However, as we demonstrate empirically in this paper, the accuracy of neural network models can vary by up to 68.1\% based only on the deep learning framework being used (PyTorch~\citep{NEURIPS2019_bdbca288} or TensorFlow~\citep{abadi2016tensorflow}) even when they share the same neural network architecture, optimization algorithms and choice of training hyperparameters. This problem is further exacerbated if other parameters are considered for optimizing performance (e.g., hardware platform, mixed precision training, network compression, etc.)~\citep{thompson2020computational}. 

To address these challenges and enable the community to build reproducible implementation of neural network models across diverse hardware and software platforms, we developed the CrossedWires dataset that allows for  a systematic comparision of PyTorch and Tensorflow models. The creation of the dataset  will democratize the study of such framework-based semantic differences between neural network models despite syntactically identical network architectures, hyperparameters and optimization algorithms. The contributions from our dataset can be summarized as follows:

\begin{itemize} 
\item The CrossedWires dataset provides 1200 pairs of neural network models with syntactically identical architectures, training hyperparameters and optimization algorithms but different test accuracies on the CIFAR-10 dataset using three popular computer vision architectures, namely - ResNet50~\citep{he2016deep}, VGG16~\citep{simonyan2014very} and DenseNet121~\citep{huang2017densely}. The accuracies of models differ by up to 68.1\%, 45.2\% and 45.7\% for VGG16, ResNet50 and DenseNet121 respectively.
    \item The CrossedWires dataset is made accessible using a Python API interface that allows the end user to load a specific model and the metadata about the accuracy of the models as a Python dataframe and a metadata data sheet (see Supplementary Material). This enables end users to analyze models with divergent behaviors without explicitly downloading the 340GB data that describes all the weights. Our dataset allows for extreme ease of use across the board, needing no more than a few lines of code from installation to neural network initialization with pre-trained weights. In this way, users can access everything from weight and bias vectors, additional model evaluation, and continued training and transfer learning. 
    \item  CrossedWires is also a \emph{living} dataset, where by other users can contribute data and models to study semantic differences between deep learning models. Especially in the context of parameter-rich models (e.g., GPT-3 with 175 billion parameters), such an effort would be needed to reconcile differences and improve performance on various `real-world' tasks. 
\end{itemize}

We feel that the CrossedWires dataset will be useful across multiple fronts. One, for researchers interested in  improving reproducibility of deep learning models, the dataset provides a starting point for testing such approaches without the need to extensively train identical networks. With the ability to extend the datasets through a functional API, it will facilitate interactions across disciplines, breaking barriers of communication such that models may be implemented in a more standardized manner, irrespective of hardware/software platforms. Two, it will also enable engineers of novel hardware and software platforms to rigorously test assumptions about model equivalence, starting with the fundamental building blocks of neural networks (e.g., densely connected layers, convolutional layers, etc.). Complementary to community-led efforts such as MLPerf~\citep{mlperf}, where benchmarking performance across software and hardware platforms have been the primary goal, CrossedWires will enable research in understanding how variability can arise and spur novel research in developing effective interfaces between software and hardware platforms with the primary goal of reproducibility. This will also enable researchers in the area of safety critical systems to quantitatively evaluate and address limitations of deploying models for various application domains. In the context of scientific applications, where surrogate models are being used to accelerate simulation kernels, the variation in model performance as a consequence of hardware and software selections will be valuable to build realistic metrics of confidence in how such models function. Inspired by the approaches developed for  scientific simulation toolkits such as molecular dynamics (MD)~\citep{openmm}, where simulations instantiated from different MD engines still provide consistent results, CrossedWires will spur the development of better interoperability across software and hardware platforms.   

\section{Methods}

\subsection{Models, Dataset, and Training Details}

The CrossedWires dataset comprises of models from hyperparameter optimization runs on three different CNN architectures, namely  ResNet50, VGG16 and DenseNet121 that are trained on the CIFAR10 dataset. The primary goal of these searches is to explore the difference in accuracy of syntactically identical models and create a survey-type global overview of the hyperparameter space. Each search is orchestrated using our Python package SpaceRay, resulting in the following artifacts: (i) 2400 total models (1200 each of PyTorch and TensorFlow), (ii) CSV files generated by \texttt{RayTune}~\citep{liaw2018tune} on the metrics of each trial, and (iii) pickled Python files containing \texttt{scipy}~\citep{2020SciPy-NMeth} \texttt{OptimizeResult} objects which detail the optimization history of the Gaussian process guided search. All the models and the metadata generated during the search process are included in the CrossedWires dataset.

Ease of use was a key design consideration in developing the dataset. Without the ability to integrate into existing codebase or a simple way to retrieve a trained network which works ``out of the box'', potential insights could be lost. All objects, including all trained TensorFlow and PyTorch models, may be accessed using the Python API which makes up CrossedWires module. Details on installation and interaction may be found as part of the repository \url{https://github.com/maxzvyagin/crossedwires} and an associated comprehensive documentation site at \url{https://crossedwires.readthedocs.io}. The entire dataset can also be downloaded as a single archive \url{https://storage.googleapis.com/crossed-wires-dataset/full_cifar10_results.zip}. However, the majority of users will likely benefit more from programmatic access versus an \emph{en masse} download.

Our experiments focused on the characteristic deep learning problem of image classification, using three well-known convolution neural network architectures: VGG16 \citep{simonyan2014very}, ResNet50 \citep{he2016deep}, and DenseNet121 \citep{huang2017densely}. All models were loaded from pre-existing definitions as part of the PyTorch and TensorFlow libraries, with default weight initialization. Importantly, the PyTorch library version of VGG16 had an extra layer of adaptive pooling included to stabilize the gradients which added approximately an extra 100 million parameters when compared to the TensorFlow version. \footnote{\url{https://pytorch.org/vision/stable/_modules/torchvision/models/vgg.html\#vgg16}} This layer was removed in order to keep the network definitions consistent -- the updated PyTorch model definition is included in our repository.  The library defined models were selected due to their likelihood of use in common machine learning scenarios, in contrast to potential users writing the code for the entire network from scratch.

The well-known dataset CIFAR10 \citep{cifar10} was selected for this benchmarking experiment, allowing for a moderate level of challenge while simultaneously not requiring immense computational resources. The standard training and test splits were loaded using the PyTorch and TensorFlow libraries, which included 50,000 training samples and 10,000 testing samples. Shuffling was turned off and data values normalized between the values of 0 and 1. Categorical cross entropy as used as the loss function, and the random seed was set to 0. The model implementations and training scripts are available at \url{https://github.com/maxzvyagin/cross_framework_hpo}.

\subsection{Hyperparameter Optimization}
\begin{wraptable}{r}{5.5cm}  \caption{Hyperparameter bounds for HyperSpace search algorithm}
  \label{hyperparameters}
  \centering
  \begin{tabular}{ccc}
    \toprule
    Hyperparameter     & Min     & Max \\
    \midrule
    Learning Rate & $1e^{-8}$ & $0.1$     \\
    Epochs  & $1$ & $50$    \\
    Adam$\epsilon$ & $1e^{-8}$  & $1.0$  \\
    Batch Size & $10$  & $10,000$  \\
    \bottomrule
  \end{tabular}
\end{wraptable} In order to effectively expose the largest  divergence of the two different deep learning frameworks, a methodical hyperparameter optimization search strategy was used. The objective function used in the optimization attempted to maximize the absolute value of the difference between the PyTorch and TensorFlow test set accuracy metrics. Four different hyperparameters were tuned - learning rate, batch size, total training epochs, and the epsilon parameter in the Adam optimizer. A relatively wide range for these parameters was fed into the HPO algorithm in order to ensure an exhaustive search. These bounds are shown in Table \ref{hyperparameters}. 

We used the approach outlined in the HyperSpace hyperparameter optimization library to examine the optimal settings for the various deep learning networks. The experiments reported here tuned the 4 hyperparameters, leading to ($2^4 = 16$)  hyperparameter spaces being fed into \texttt{scikit-optimize}, with an overlap factor $\phi=0.5$. Our implementation is a more automated and user friendly extension of the hyperparameter search defined in the original HyperSpace implementation \citep{hyperspace}. As demonstrated previously, HyperSpace can discover hyperparameter settings that can outperform other approaches and result in optimal settings for a model's performance \citep{young2018hyperspace, young2020distributed}. 

\begin{figure}[ht]
\includegraphics[width=\linewidth]{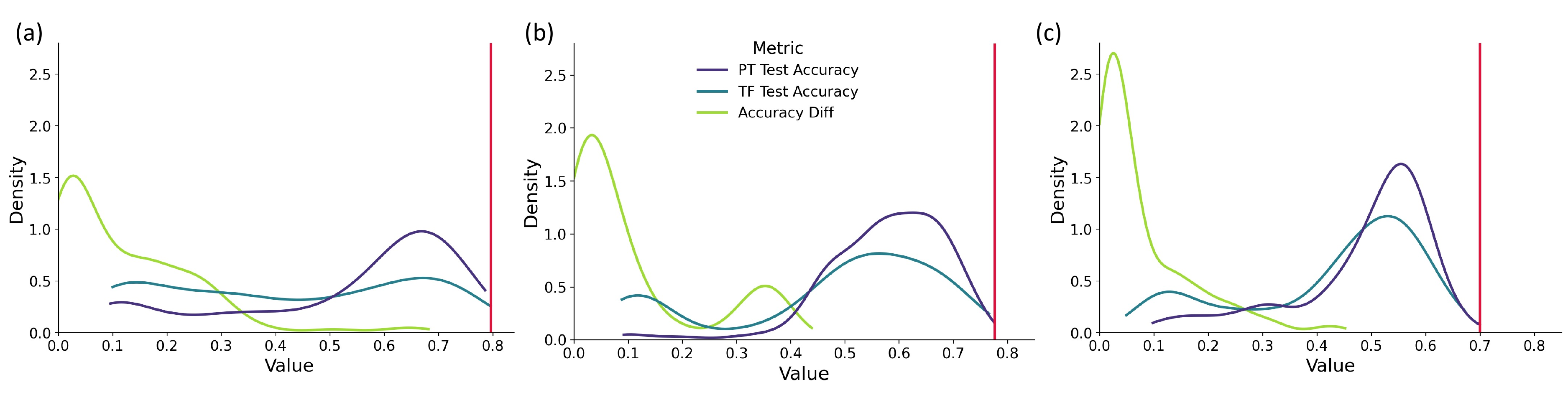}
\caption{Kernel density estimation plots to showcase distribution of individual framework accuracy and accuracy divergence values across each architecture's 400 HPO trials. The red line in each plot corresponds the maximum test set accuracy for the corresponding architecture. (a) corresponds to VGG16, (b) corresponds to ResNet50, and (c) corresponds to DenseNet121.}
\label{fig:kde_plots}
\end{figure}
These individual spaces are then fed into the \texttt{Ray Tune} \texttt{SkOptSearch} implementation, which integrates with the \texttt{scikit-optimize} Gaussian process search algorithm to optimize within each hyperparameter space and evaluates them in parallel. This entire process is handled by the SpaceRay library (\url{www.github.com/maxzvyagin/spaceray}). For each generated \texttt{scikit-optimize} search space, 25 trials were performed, for a total of 400 tested hyperparameter sets per neural network architecture. For each architecture, this then translates to 800 sets of model weights (400 PyTorch, 400 TensorFlow), for a total of 2400 model weights and 1200 tested hyperparameter configurations. All trials were logged using Weights and Biases \citep{wandb} and the links to these logs are included in the documentation and the supplementary material. In addition, the CSV file exports of the Weights and Biases logs are directly included as a part of the CrossedWires dataset.

\subsection{Hardware and Environment Details}
The exploratory runs as well as the dataset generation experiments were performed on the Lambda system at Argonne National Laboratory. Each node in the Lambda system has 8 NVIDIA Tesla V100 GPUs and 80 Intel Xeon Gold 6148 CPUs. The system utilized NVIDIA driver version 470.57.02, and CUDA version 11.4. Each trial was automatically allocated a single GPU and CPU by RayTune. While replication of the dataset will only require approximately $\mathcal{O}(10^2)$ GPU-hours, our exploratory studies to determine the sensitive hyperparameters and parameter regimes took $\mathcal{O}(10^4-10^5)$ GPU-hours.

\section{Benchmark Results and Dataset Observations}

\subsection{Distribution of Results}
As expected, there is a wide range in the final accuracy metrics for both TensorFlow and PyTorch, in addition to a surprisingly wide range of total divergence between the two frameworks, depending on the selected hyperparameter settings. The distributions of the individual accuracy and accuracy difference metrics are displayed in Figure \ref{fig:kde_plots}. Overall, we see that the majority of the HPO trials generate high test set accuracy in both frameworks, despite the accuracy distribution not being exactly the same. Noticeably, PyTorch has a greater overall density at the top end of the accuracy spectrum across all three architectures. In addition, we see that the majority of the accuracy difference distributions (green line) are at or around zero, with a markedly higher density of non-zero trials in the DenseNet architecture (Figure \ref{fig:kde_plots}c). From these plots, it is clear that all three HPO searches contain trials with a non-trivial accuracy difference between PyTorch and TensorFlow models of above 5\%. Overall, 52.3\% of the trials that make up the dataset resulted in a model pair which diverges by at least 5\%.

\begin{table}[h]
  \caption{Statistical summary of the accuracy difference metric for each tested architecture.}
  \label{tab: acc_diff_summary}
  \centering
  \begin{tabular}{ccccc}
    \toprule
    Architecture     & Min  & Max & Median & Mean \\
    \midrule
    VGG & $4.17*e^{-10}$ & $0.681$ & $0.0843$ & $0.123$   \\
    ResNet & $1.36*e^{-4}$ & $0.452$ & $0.0418$ & $0.0779$ \\
    DenseNet & $2.17*e^{-4}$ & $0.457$ & $0.0563$ & $0.104$ \\
    \bottomrule
  \end{tabular}
\end{table}

Within each architecture, there are a range of hyperparameter configurations which lead to high accuracy in both frameworks, those which lead to low accuracy in both frameworks, those which lead to high PyTorch accuracy but low TensorFlow accuracy, and vice versa. Unsurprisingly, the final accuracy statistics, including divergence between the frameworks, varies based on the network architecture. The range of the test set accuracy for each individual framework does not necessarily have a linear relationship with the number of parameters within the network.
Interestingly, the lowest TensorFlow accuracy in DenseNet is zero, while it is non-zero for the other two architectures (Table \ref{tab: framework_acc_summary}).

\begin{table}[h]
  \caption{Statistical summary of individual framework accuracy on the test dataset, for both PyTorch (PT) and TensorFlow (TF).}
  \label{tab: framework_acc_summary}
  \centering
  \begin{tabular}{ccccc}
    \toprule
    Architecture & PT Min  & TF Min & PT Max & TF Max \\
    \midrule
    VGG & $0.0959$ & $0.100$ & $0.786$ & $0.796$   \\
    ResNet & $0.0978$ & $0.0493$ & $0.699$ & $0.671$ \\
    DenseNet & $0.0799$ & $0.0$ & $0.764$ & $0.771$ \\
    \bottomrule
  \end{tabular}
\end{table}
\begin{table}[h]
\centering
\caption{Top three and bottom three sets of hyperparamter results for VGG architecture when sorted by accuracy difference.}
\label{tab:top_and_bottom_vgg}
\resizebox{\textwidth}{!}{
\begin{tabular}{ccccccc}
\toprule
 Adam Epsilon &  Batch Size &  Learning Rate &  Epochs &  PT Test Acc &  TF Test Acc &  Accuracy Diff \\
\midrule
      0.19658 &         202 &        0.09901 &      50 &      0.09998 &      0.78150 &        0.68152 \\
      0.73954 &          22 &        0.06213 &      21 &      0.75619 &      0.10000 &        0.65619 \\
      0.16867 &         242 &        0.09234 &      21 &      0.75329 &      0.10000 &        0.65329 \\
\midrule
      0.84102 &          10 &        0.05205 &      21 &      0.10000 &      0.10000 &        0.00000 \\
      0.07937 &          10 &        0.04891 &       7 &      0.10000 &      0.10000 &        0.00000 \\
      0.35535 &          10 &        0.04866 &      46 &      0.10000 &      0.10000 &        0.00000 \\
\bottomrule
\end{tabular}
}
\end{table}

Overall, the VGG architecture has the greatest level of maximum model divergence at 68.1\%, with the maximum in ResNet and DenseNet architectures being about 23\% less (Table \ref{tab: acc_diff_summary}). Notably, although the individual framework statistics are similar for VGG and DenseNet, the maximum accuracy divergence is 20\% higher in VGG. In other words, the top trials of the VGG architecture results in Table \ref{tab:top_and_bottom_vgg} generate models where one framework reaches a performance level similar to the individual maximums seen in Table \ref{tab: framework_acc_summary}. However, in the DenseNet architecture results in Table \ref{tab:top_and_bottom_densenet}, even the high performing network in the top trials falls short of the 76-77\% maximum performance level (Table \ref{tab: framework_acc_summary}). This is a key result, as it is then plausible that a top performing set of hyperparameters for VGG (selected using a single framework search as is normally done) could be unstable in the other framework. In DenseNet, on the other hand, this is not the case, as the top ``divergent'' parameter configurations do not lead to maximum performance in either framework. The same holds true for the ResNet architecture, with the individual framework accuracies in Table \ref{tab:top_and_bottom_resnet} showing a maximum potential accuracy of 61\% in an individual framework, while in Table \ref{tab: framework_acc_summary} the maximum potential accuracy is 69\%. All trial results for the three architectures can be found in the supplementary material and as part of the hosted dataset.

\begin{table}[h]
\centering
\caption{Top three and bottom three sets of hyperparameter results for ResNet architecture when sorted by accuracy difference.}
\label{tab:top_and_bottom_resnet}
\resizebox{\textwidth}{!}{
\begin{tabular}{ccccccc}
\toprule
 Adam Epsilon &  Batch Size &  Learning Rate &  Epochs &  PT Test Acc &  TF Test Acc &  Accuracy Diff \\
\midrule
      0.20390 &         382 &        0.05596 &      26 &      0.59364 &      0.09990 &        0.49374 \\
      0.20259 &         466 &        0.05103 &      24 &      0.61146 &      0.12910 &        0.48236 \\
      0.46381 &         502 &        0.07989 &      20 &      0.57465 &      0.10610 &        0.46855 \\
\midrule
      0.53622 &         904 &        0.03705 &      26 &      0.47656 &      0.47680 &        0.00024 \\
      0.18580 &         388 &        0.01245 &      20 &      0.53219 &      0.53240 &        0.00021 \\
      0.91122 &         904 &        0.07455 &      45 &      0.53747 &      0.53730 &        0.00017 \\
\bottomrule
\end{tabular}
}
\end{table}

\begin{table}
\centering
\caption{Top three and bottom three sets of hyperparameter results for DenseNet architecture when sorted by accuracy difference.}
\label{tab:top_and_bottom_densenet}
\resizebox{\textwidth}{!}{
\begin{tabular}{ccccccc}
\toprule
 Adam Epsilon &  Batch Size &  Learning Rate &  Epochs &  PT Test Acc &  TF Test Acc &  Accuracy Diff \\
\midrule
      0.37500 &         589 &        0.05955 &       2 &      0.54946 &      0.11050 &        0.43896 \\
      0.62937 &         824 &        0.08118 &       2 &      0.52159 &      0.08830 &        0.43329 \\
      0.37500 &         686 &        0.06250 &       3 &      0.56427 &      0.15010 &        0.41417 \\
\midrule
      0.08772 &         602 &        0.02302 &      46 &      0.73392 &      0.73320 &        0.00072 \\
      0.71086 &         370 &        0.08254 &      43 &      0.71602 &      0.71650 &        0.00048 \\
      0.60260 &          13 &        0.05400 &       1 &      0.09990 &      0.10000 &        0.00010 \\
\bottomrule
\end{tabular}
}
\end{table}

\begin{figure}
\includegraphics[width=\textwidth]{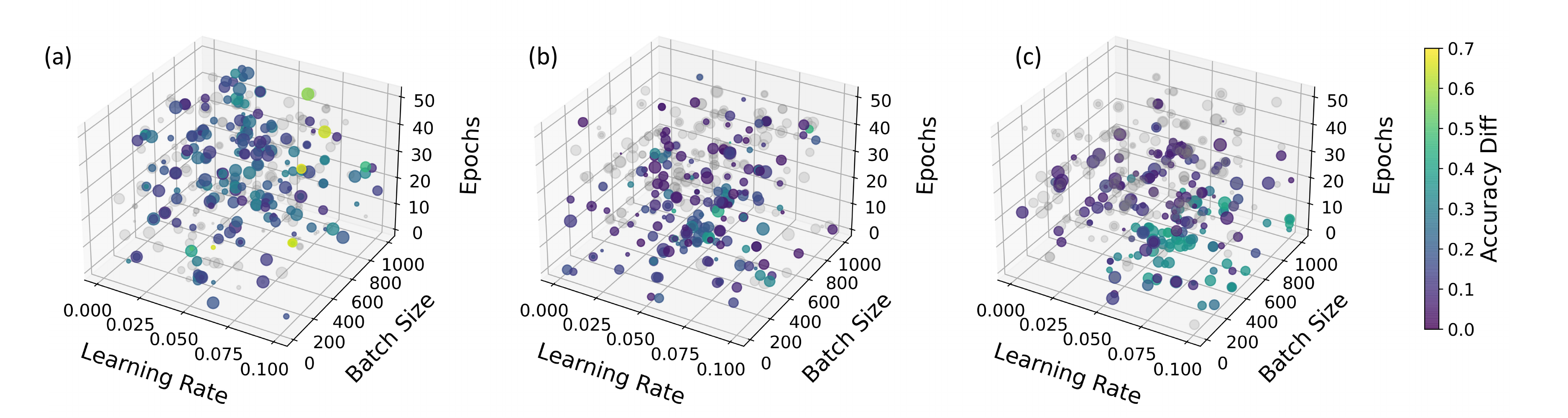}
\caption{Visualization of hyperparameter space with respect to accuracy difference between PyTorch and TensorFlow. Size of points corresponds to epsilon parameter of Adam. Points are painted if above threshold of median accuracy difference. Median statistics can be found in Table \ref{tab: acc_diff_summary}. (a) corresponds to VGG16, (b) corresponds to ResNet50, and (c) corresponds to DenseNet121.}
\label{fig:corr_plots}
\end{figure}

\subsection{Hyperparameter Space Topology}

\begin{figure}[h]
    \centering
    \includegraphics[scale=0.35]{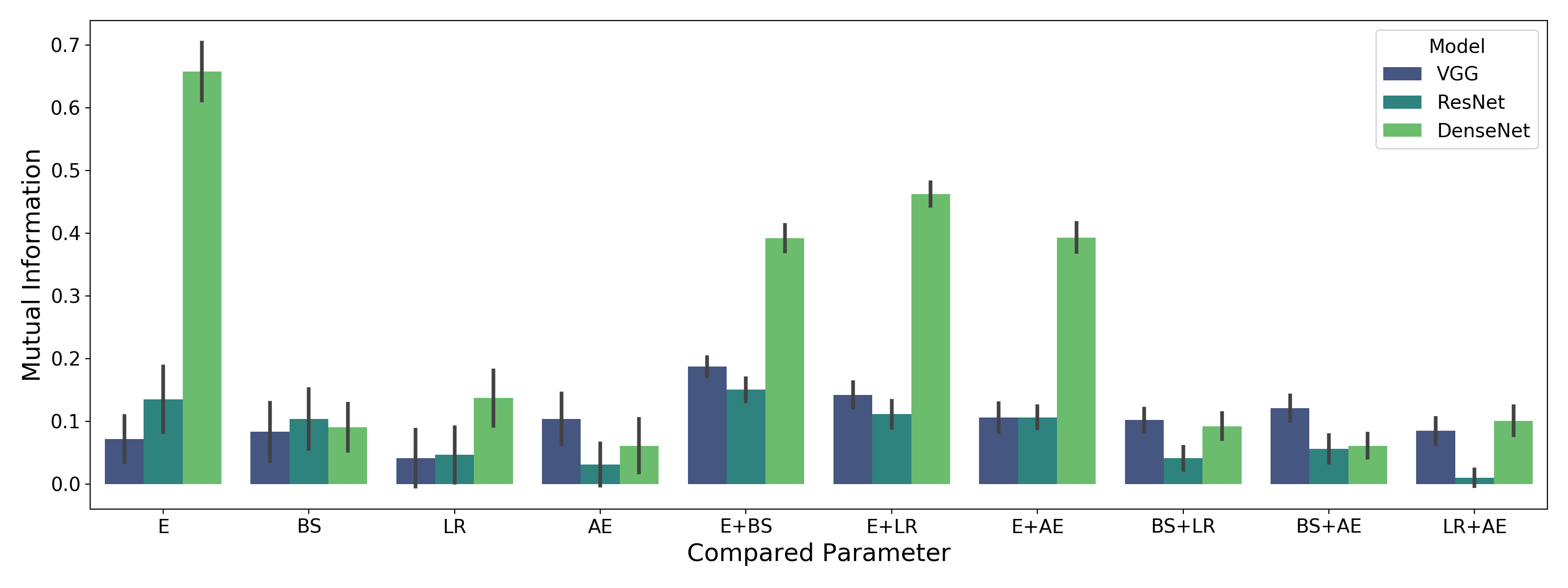}
    \caption{Mutual information between hyperparameter settings (individual and pairwise) and accuracy difference. All hyperparameter settings and mutual information values were normalized between 0 and 1, then digitized with bin sizes of 0.1 \citep{harris2020array}. Univariate and bivariate values have been calculated. Both orders of the bivariate data were computed, with the average taken as the final result. To address the non-deterministic nature of this calculation, the computation was repeated 50 times - the mean and standard deviation are shown here. Mutual information was calculated using the NPEET library~\citep{ver-steeg}.}
    \label{fig:mutual_information}
\end{figure}

Through Figures \ref{fig:corr_plots} and \ref{fig:mutual_information}, it is clear that the hyperparameter configurations leading to divergent results between PyTorch and Tensorflow are not homogeneous across the different network architectures. In Figure \ref{fig:corr_plots}, the 4D hyperparameter space is visualized against the top 50\% of accuracy difference results (bottom 50\% are colored in gray), painted by the level of accuracy difference. The distribution of these points appears to be unique to each individual architecture without any clear regions of similarity across the three architectures. However, there does appear to be some clustering: for example, in the DenseNet figure \ref{fig:corr_plots}c, there is a clear cluster in the very center of the axes of an $\sim0.4$ accuracy difference, with turquoise colored points. The greatest variation in the hyperparameter topology of the painted points does seem to be in the VGG model results (Figure \ref{fig:corr_plots}a) - there are some small clusters present, however, the majority of the points are distributed somewhat evenly across the entire visualized hyperparameter space. 

By calculating the mutual information between the hyperparameter settings and the final accuracy difference between the two frameworks, it is possible to infer the importance and impact of each individual hyperparameter setting on divergence of the resulting models. Both univariate and bivariate results were computed, to understand the effect of both individual and pairwise sets of the hyperparameters. This analysis is presented in Figure \ref{fig:mutual_information}. From this, we see that the Adam epsilon parameter has the highest influence on the final accuracy difference in the VGG architecture, the pair of epochs and Adam epsilon has the highest influence in ResNet, and the epochs parameters has the highest influence in DenseNet. This is a key understanding: it shows that architecture must be considered when attempting to create stable, framework agnostic hyperparameter settings. This plot also demonstrates that different categories of hyperparameter settings must not be discounted when a novel architecture is involved. For example, the Adam epsilon parameter is not of particularly high importance in the DenseNet architecture, but it is the maximally important parameter in VGG. If this hyperparameter search-based experiment was only performed on DenseNet, with the intent to extrapolate these results, one could erroneously conclude that the epsilon value could instead be hardcoded in hyperparameter searches for subsequent architectures such as VGG. Further, the relatively high values of mutual information found within the bivariate computations show that hyperparamaters should not only be considered in isolation. 

Another advantage of using our hyperparameter optimization strategy is that it enables one to examine the partial dependency of the hyperparameters and its impact on model performance (see supplementary material, partial dependency plots). Given that very few settings of hyperparameters actually end up `sampling' good model performance (i.e.,  accuracy), understanding the innate structure of the hyperparameter space can be valuable in fine tuning the models across different frameworks. While this represents just a snapshot of the 4-dimensional landscape (spanned by the hyperparameters), complex models can include higher-order dependencies that may need principled approaches to characterize the overall performance space of semantic equivalence. Such insights are enabled within our dataset implicitly via our data API and SpaceRay.

\section{Related Work}

\paragraph{Machine learning benchmarks} Benchmarks are a key area of interest to the machine learning community, allowing for effective tracking, collective knowledge, and state of the art performance. An example of this kind of collection is MLPerf, a benchmarking suite which is used to measure and record the performance of deep learning networks on various hardware systems \citep{mlperf}. This organization of knowledge is key in keeping the industry moving forward in pursuit of continuously accelerating training and inference times. Another example is the Papers with Code website, which aggregates accuracy metrics on different architectures and datasets, showcasing which submissions are currently state of the art and linking the statistics to the relevant papers and repositories \citep{PWC}. However, to our knowledge there is  very little data in understanding how model performance can vary as a consequence of the choice of a particular framework, hyperparameter settings and how the model is consequently optimized and deployed. %

In this paper, we focus on two of the most popular deep learning frameworks, PyTorch and TensorFlow. Both frameworks implement the sets of kernels, layers, and operations required to realize neural networks, however, they do so in significantly different ways. PyTorch is imperative and based on dynamic computation \citep{NEURIPS2019_bdbca288}. TensorFlow, on the other hand, on is graph-based, allows for kernel subsitution at runtime, and is automatically tuned for specific hardware based on resource allocation \citep{abadi2016tensorflow}. There are also noted differences in certain mathematical operation implementations such as the convolution and optimization algorithms. Both of them leverage the cuDNN library \citep{chetlur2014cudnn}, a set of primitives used to accelerate graphics processing unit (GPU) computation. Peer reviewed sources do exist which compare  hardware performance between PyTorch and TensorFlow, such as in \cite{performance-char-dnn} which performed a review of deep networks on modern CPU clusters.

\paragraph{Common Representation of machine learning models }At this time, translation between deep learning frameworks can be cumbersome. One potential solution to this has been the introduction of the Open Neural Network Exchange (ONNX) \citep{bai2019}, which attempts to address framework translation by defining a separate intermediate representation. Although widely adopted for its ease of representation, translating larger and more complex deep learning models and specific aspects (perhaps unique to an implementation framework or hardware) can be challenging to capture within its description. Other problems include translation of the architecture and layers which make up the network, and the translation of operations such as custom training loops. This also does not account for the potential that inference mechanisms could vary between frameworks. Often, the path of least resistance is to simply re-train in a new framework, however, this comes with its own challenges, of which hyperparameter optimization is just one. We posit that by co-training two syntactically equivalent networks across different implementation frameworks, one can obtain valuable insights into the nature of affects that hyperparameters may induce on model performance within a unified evaluation framework. The CrossWire dataset provides a starting point for such evaluations.

\section{Conclusions}

In this paper, we have presented the initial generation and statistical characterization of a dataset which will allow for improved understanding of the differences between deep learning framework implementations. Our dataset could prove to be highly useful to better understand the dependence of model implementation in the context of complex hyperparameter spaces which eventually plays an essential role in its deployment and successful application. By providing this dataset to the machine learning community, we hope to provide researchers with the key to unlock novel approaches that improve the stability of machine learning models, and more importantly to develop a rational understanding of how to approach hyperparameter settings. Ideally, this dataset will allow for generalizable insights that will make way for informed and stable hyperparameter configuration without requiring extreme levels of compute power and extensive hyperparameter searches. 

 It is important to note that while this dataset only consists of the three models as applied to the CIFAR-10 dataset, a significant number of other models (e.g., multi-layer perceptron, Alexnet \citep{krizhevsky2012imagenet}, etc.) and datasets were explored (including Fashion-MNIST \citep{xiao2017fashion}, ImageNet \citep{deng2009imagenet}, etc.) were explored. In addition, the approach presented here is not limited to only computer vision models and can be extended to  complex models (for e.g., language models such as BERT \citep{devlin2018bert}, GPT \citep{radford2019language}) and other scientific datasets. Further, in the context of exploring complex hyperparameter spaces, we deliberately chose to examine a 4-dimensional space. This was a practical choice, rather than a limitation of the hyperparameter optimization approach since larger space exploration would necessarily need extensive computational resources as well. Given the opportunity to expand this dataset, we will gladly incorporate larger models and hyperparmeter spaces. We hope that as a living dataset, our efforts will continue to evolve and involve the broader scientific community in contributing to our dataset and enable better development of reproducible deep learning models. This also reflects on how important it would be to quantify the inter- and intra-framework variability when implementing foundational models for AI~\citep{bommasani2021opportunities}.
 
 This contribution to the machine learning community must be a living dataset in order to realize its full potential and provide the greatest level of utility. We invite submissions to the dataset, with the hope of expanding our knowledge base to other machine learning frameworks, datasets, architectures, and hardware systems. Submissions must be formatted and include all code in order to be considered as a contribution, in addition to following standard reproducibility guidelines. Details for contribution are included in the documentation, and will take the form of a GitHub pull request which maintains the streamlined API structure. %

\begin{ack}
Funding for this work was provided by the Department of Energy’s Advanced Scientific Computing Research Program through grant number 31975.2 to Argonne National Laboratory for the Co-Design of Advanced Artificial Intelligence (AI) Systems for Predicting Behavior of Complex Systems Using Multimodal Datasets project. The research was partially supported by the National Science Foundation SPX award \#1822976, National Science Foundation award \#2113307, DARPA GARD contract \#HR00112020002, and ONR Science of AI program grant \#N00014-21-1-2332. This research used resources of the Argonne Leadership Computing Facility, which is a DOE Office of Science User Facility supported under Contract DE-AC02-06CH11357.
\end{ack}

\bibliographystyle{plainnat}
\bibliography{references}

\end{document}